# The Real-World-Weight Cross-Entropy Loss Function: Modeling the Costs of Mislabeling

**Yaoshiang Ho, Samuel Wookey**
Thinky.AI Research, Los Angeles, CA 90027 USA

Corresponding author: Yaoshiang Ho (email:yaoshiang@thinky.ai)

**ABSTRACT** In this paper, we propose a new metric to measure goodness-of-fit for classifiers: the Real World Cost function. This metric factors in information about a real world problem, such as financial impact, that other measures like accuracy or F1 do not. This metric is also more directly interpretable for users. To optimize for this metric, we introduce the Real-World-Weight Cross-Entropy loss function, in both binary classification and single-label multiclass classification variants. Both variants allow direct input of real world costs as weights. For single-label, multiclass classification, our loss function also allows direct penalization of probabilistic false positives, weighted by label, during the training of a machine learning model. We compare the design of our loss function to the binary cross-entropy and categorical cross-entropy functions, as well as their weighted variants, to discuss the potential for improvement in handling a variety of known shortcomings of machine learning, ranging from imbalanced classes to medical diagnostic error to reinforcement of social bias. We create scenarios that emulate those issues using the MNIST data set and demonstrate empirical results of our new loss function. Finally, we discuss our intuition about why this approach works and sketch a proof based on Maximum Likelihood Estimation.

**INDEX TERMS** machine learning, class imbalance, oversampling, undersampling, ethnic stereotypes, social bias, maximum likelihood estimation, cross-entropy, softmax

## I. INTRODUCTION

Over the past several years, deep learning has achieved dramatic success in areas ranging from image recognition to speech recognition and decision making. As deep learning has been rapidly commercialized, issues have arisen.

First, machine learning has shortcomings when analyzing imbalanced classes [1]. A simple and well-known thought experiment motivates the problem: when seeking to identify an uncommon disease with say a 1% occurrence, a trivial classifier that never predicts that disease is 99% accurate. Measures such as recall, precision, and F1 score are sometimes used as alternative measures of goodness-of-fit. Practitioners have also introduced heuristics like oversampling, undersampling, and weighted labels [2].

Second, the approach of using softmax and its associated loss function for single-label, multiclass classification means that probabilistic false negatives are not directly penalized during training. This limits the ability to solve a variety of issues, ranging from reinforcement of social bias [3] to medical diagnostic errors that deserve extra scrutiny [4].

In this paper, we propose a new metric for classifiers: the Real World Cost. This metric is more comprehensive than accuracy or F1 because it applies weights to each error based on an estimate of its impact in the real world, such as financial cost. The result is a metric that will be more familiar for some end users, compared to abstract concepts like F1 or overly simplistic measures like accuracy.

We also describe and build an efficient implementation of a new loss function we call the "Real-World-Weight Cross-Entropy" (RWWCE), which is designed to optimize for the Real World Cost. We find that RWWCE is a generalization of binary cross-entropy and softmax cross-entropy (which is also called categorical cross-entropy). Specifically, RWWCE adds weights to address false positives.

Furthermore, we introduce a framework to set these weights based on factors that exist in the dataset or underlying problem to be solved. The framework is grounded in underlying costs of the real world problem, which means they should be discovered and set only once. These weights are not a machine learning model's hyperparameters, which would require repeated adjustments based on heuristics.

We apply the RWWCE loss function against binary imbalanced data to demonstrate improvements in Real World Cost. Our control group includes the approach of training a binary classifier neural network for accuracy, but then after training, tuning one of the hyperparameters (threshold) to optimize F1. We then discuss the similarities and differences between the RWWCE loss function and the heuristics of oversampling, undersampling, and weighted labels.

Subsequently, we apply our RWWCE loss function towards single-label, multiclass classification, activated by softmax. We choose specific combinations of target label and possible incorrect predicted label to represent high expense or socially unacceptable mistakes. We demonstrate a reduction in these specific mislabeling errors and a reduction in Real World Cost. We also analyze the design of the softmax cross-entropy loss function to identify the specific additional capabilities of the RWWCE loss function.

Finally, we sketch the outline of a proof, based on the underlying concepts of Maximum Likelihood Estimation, and conclude with future directions.

## II. RELATED WORK

Widely available machine learning libraries like TensorFlow support weighting of the loss function [5]. For binary classification, the binary cross-entropy loss function can have a weight applied to the probabilistic false negative case. Setting this value greater than one increases the penalty for probabilistic false negatives during training. The value can be set to less than one to decrease the penalty for probabilistic false negatives. For multi-class classification, the categorical cross-entropy loss function can be weighted by class, increasing or decreasing the relative penalty of a probabilistic false negative for an individual class. Class-Balanced Loss sets these weights in proportion to the inverse of the number of samples per class [6].

Focal Loss sets weights based on class and difficulty of classification. Easy to classify examples are given less weight [7].

The idea of adding weights to increase the cost of specific combinations of targeted and predicted class in single-label, multiclass classification is described in [8].

Oversampling synthetically creates additional minority examples by replicating data points from a minority class. Undersampling removes some amount training data from the majority class. This has the added benefit of reducing the size of the training data. Both techniques and their extensions are surveyed in [9].

The Synthetic Minority Over-sampling Technique (SMOTE) creates synthetic training data in a more sophisticated way than plain oversampling [10]. Applied to a minority class, it synthesizes additional training data by interpolating between existing data points.

Techniques have been developed to reduce social bias in some neural networks designs such as adjusting the high dimensional vectors representing individual words to remove differences in the distance from the concepts of male and female in word2vec [11] and the Seldonian approach of describing and regulating undesirable behavior [12].

For pairs of a false negative and false positive that reference racist tropes, a high profile problem was reduced by eliminating the label entirely, an extreme version of a more general practice of ignoring the outputs from a machine learning model when confidence is below a certain threshold [13].

Weighted maximum likelihood estimators [14] address the challenges of imbalanced classes from the perspective of Maximum Likelihood Estimation [15]. We address this further in the sketch of the proof of our RWWCE.

In the case of binary classification, F1 is a common measure of goodness-of-fit for imbalanced classes. Weighted Maximum Likelihood was applied to optimize F1 score in [16], and, loss functions have been developed to allow a machine learning model to directly optimize F1 score [17].

## III. REAL-WORLD-WEIGHT CROSS-ENTROPY LOSS FUNCTION

During neural network training, the cost function is the key to adjusting a neural network's weights to create a better fitting machine learning model. Specifically, during forward propagation, the neural network is run on training set data, and outputs are generated which in the case of classification indicate the probability or confidence in possible labels. These probabilities are compared to the target labels, and, the loss function calculates a penalty for any deviation between the target label and the neural network's outputs. During backpropagation the partial derivative of the loss function is calculated for each trainable weight of the neural network. The weights are adjusted by these partial derivatives. Under normal conditions, backpropagation iteratively adjusts the trainable weights of a neural network to produce a model with lower loss.

The standard binary cross-entropy loss function is given by:

$$J_{bce} = -\frac{1}{M}\sum_{m=1}^{M}[y_m \times \log(h_\theta(x_m)) + (1-y_m) \times \log(1-h_\theta(x_m))] \quad (1)$$

where
- $M$    number of training examples
- $y_m$    target label for training example m
- $x_m$    input for training example m
- $h_\theta$    model with neural network weights $\theta$

The first term, $y_m \times \log(h_\theta(x_m))$, disincentivizes probabilistic false negatives during training. For example, suppose a training example has target 1, the output of the machine learning model is 0.6. We say that there is a probabilistic false negative of 40%. In other words, from a Bayesian perspective, the model has 40% confidence in the wrong result. Or from a Frequentist perspective, the model will be wrong 40% of the time. The loss function penalizes this 40% by returning the value -log(0.6) = 0.22. In the perfect case, if the binary classifier outputs 1, then it is completely accurate for the training example and the loss is

-log(1) = 0. The same logic applies for the second term and probabilistic false positives.

The standard weighted binary cross-entropy loss function is given by:

$$J_{wbce} = -\frac{1}{M}\sum_{m=1}^{M}[w \times y_m \times \log(h_\theta(x_m)) + (1-y_m) \times \log(1-h_\theta(x_m))] \quad (2)$$

where
- $M$    number of training examples
- $w$    weight
- $y_m$    target label for training example m
- $x_m$    input for training example m
- $h_\theta$    model with neural network weights $\theta$

The additional weight can be set to adjust the importance of the positive labels. A common use is to give more weight to minority classes.

For the case of single-label, categorical classification (i.e. softmax activation) the standard categorical cross-entropy loss is given by:

$$J_{cce} = -\frac{1}{M}\sum_{k=1}^{K}\sum_{m=1}^{M} y_m^k \times \log(h_\theta(x_m, k)) \quad (3)$$

where
- $M$    number of training examples
- $K$    number of classes
- $y_m^k$    target label for training example m for class k
- $x$    input for training example m
- $h_\theta$    model with neural network weights $\theta$

The standard weighted categorical cross-entropy loss is given by:

$$J_{wcce} = -\frac{1}{M}\sum_{k=1}^{K}\sum_{m=1}^{M} w_k \times y_m^k \times \log(h_\theta(x_m, k)) \quad (4)$$

where
- $M$    number of training examples
- $K$    number of classes
- $w_k$    weight for class k
- $y_m^k$    target label for training example m for class k
- $x_m$    input for training example m
- $h_\theta$    model with neural network weights $\theta$

The Real-World-Weight Cross-Entropy (RWWCE) loss function introduces weights on the cost of missing a positive, and a separate weight for missing a negative. For binary classification, RWWCE loss function is given by:

$$J_{brwwce} = -\frac{1}{M}\sum_{m=1}^{M}[w_{mcfn} \times y_m \times \log(h_\theta(x_m)) + w_{mcfp} \times (1-y_m) \times \log(1-h_\theta(x_m))] \quad (5)$$

where
- $M$    number of training examples
- $w_{mcfn}$    marginal cost of a false negative over true positive
- $w_{mcfp}$    marginal cost of a false positive over true negative
- $y_m$    target label for training example m
- $x_m$    input for training example m
- $h_\theta$    model with neural network weights $\theta$

For the single-label, categorical classification, the RWWCE is given by:

$$J_{crwwce} = -\frac{1}{M}\sum_{k=1}^{K}\sum_{m=1}^{M}\left[w_{mcfn}^k \times y_m^k \times \log(h_\theta(x_m, k)) + \sum_{k'=1}^{K} w_{mcfp}^{k,k'} \times y_m^k \right.$$
$$\left. \times \log\left(1-h_\theta(x_m, k')\right)\right]$$
$$\text{s.t.}$$
$$k' \neq k \quad (6)$$

where
- $M$    number of training examples
- $K$    number of classes
- $y_m^k$    target label for training example m for class k
- $h_\theta$    model with neural network weights $\theta$
- $x_m$    input for training example m
- $w_{mcfn}^k$    marginal cost of a false negative over a true positive
- $w_{mcfp}^{k,k'}$    marginal cost of a false positive of class k' over a true negative, when the true positive is k

The false positive matrix $w_{mcfp}$ has unused values along the main diagonal, because they represent true negatives, not false positives. This matrix and its associated triple summation in the loss function is the essence of modeling additional loss in the case of probabilistic false positives. In future work, we will explore multilabel, multiclass categorization. With k labels, the weight matrix could be as large as $2^k$ by $2^k$. Unlike RWWCE for binary and single-label, multiclass categorization, we have not developed an efficient implementation of a multilabel, multiclass RWWCE loss function.

## IV. BINARY CLASSIFICATION OF IMBALANCED CLASSES

We tested the RWWCE loss function against imbalanced classes for binary classification.

We created 100 data sets based on MNIST. MNIST is a widely used data set for neural network training consisting of 70,000 examples of images of hand written Arabic numerals along with labels (targets). Each example is a 28 by 28 pixel image, which we flattened to 784 pixels in a single vector.

For each of the 10 possible digits, there are roughly 7,000 examples.

For the first data set, we took the first 630 examples of the numeral "0" and labeled them to 1 (true). We took the 63,000 examples not labeled with the digit "0" and labeled them to 0 (false). This was our first data set. We divided up this data set into 67.5% training, 7.5% validation, and 25% test.

Our second through tenth data sets repeated the above procedure but with the second through tenth batches of 630 examples of the number "0".

We repeated the above 10 steps for the remaining digits "1" through "9" for a total of 100 data sets.

We trained 100 control neural networks and 100 experimental neural networks. Their only difference was the loss function. All were trained with 10 epochs and a batch size of 100. All took 784 inputs and then a penultimate dense layer of 10, activated by ReLU. The final layer was a dense layer of 1 with a sigmoid activation, the standard technique for binary classification. All were implemented with the same Keras library using TensorFlow as the backend. All were run on the Google Colaboratory platform. The code is available at https://github.com/yaoshiang/The-Real-World-Weight-Crossentropy-Loss-Function.

The control neural networks used the standard Keras binary cross-entropy loss function, which wraps the TensorFlow implementation. The experimental neural networks used the custom RWWCE loss function with weights described below.

We created a second set of 100 control neural networks. We took the first 100 control neural networks and without adjusting any of their weights, used the method of searching across all possible thresholds to maximize F1 [15]. In other words, instead of treating an output greater than the threshold of 0.5 as a prediction of true, we searched for a different threshold to use as the prediction of true.

For our experimental neural networks, we set the marginal cost of not identifying a positive (marginal cost of FN) at 2,000 (e.g. for the first three data sets, imagining the label "0" to represent a rare but expensive disease that if missed, costs $2,000 in future medical treatment and pain and suffering). We also set the marginal cost of a false positive at 100 (e.g. representing a $100 cost of a retest).

We define the Real World Cost as the sum of the marginal cost of a false negative multiplied by the number of false negatives and the marginal cost of a false positive multiplied by the number of false positives, divided by the total number of samples. Real World Cost represents deviation from the value of a perfect classifier.

The results are summarized below. The test model has fewer false negatives but even more false positives than either control model, leading to an increase in top-1 error. This is expected, because the real world cost of a false positive is far less than the real world cost of a false negative. Crucially, our Real World Cost measure indicates that our test would deliver lower real world costs. The p-values are calculated using the pairwise t-test.

TABLE I
RESULTS OF RWWCE LOSS FUNCTION FOR BINARY CLASSIFICATION

| Model | Mean FN | Mean FP | Mean Top-1 Error | Mean Real World Cost |
|---|---|---|---|---|
| Control 1 (n=100) | 45.4 | 12.7 | 0.37% | $5.78 |
| Control 2 (n=100) | 31.7 | 20.3 | 0.33% | $4.11 |
| Test (n=100) | 16.1 | 127.2 | 0.91% | $2.81 |
| p-value (Control 1 vs Test) | $1.2 \times 10^{-21}$ | $3.1 \times 10^{-28}$ | $3.0 \times 10^{-24}$ | $2.3 \times 10^{-18}$ |
| p-value (Control 2 vs Test) | $2.7 \times 10^{-35}$ | $2.5 \times 10^{-28}$ | $1.8 \times 10^{-25}$ | $3.1 \times 10^{-27}$ |

### A. COMPARING REAL WORLD COST TO F1

We argue that optimizing for Real World Cost can be superior to F1 based approaches.

First, training a neural network for one goal, accuracy, then performing an exhaustive search across all possible thresholds to optimize a second measure, F1, is a two step process that does not allow the weights of the neural network to adapt to the real goal, the F1 score. That said, there has been recent work to develop a loss function to directly optimize F1 and related scores [14].

Second, when a machine learning model is being applied to make decisions in the real world, the costs and benefits of real world outcomes comprise additional information to apply when training a machine learning model. The F1 score does not factor in this additional information.

Finally, the F1 score is mathematically focused on infrequent positives. In cases where positives are frequent, the F1 score can be high from a trivial classifier.

### B. COMPARING RWWCE TO HEURISTICS IN BINARY CLASSIFICATION

In the binary classification case, RWWCE is mathematically equivalent to the widely available weighted binary cross-entropy. (As we will see, in the categorical case RWWCE is more expressive than the weighted categorical cross-entropy function). RWWCE allows direct application of costs for false positives and false negatives, whereas weighted binary cross-entropy allows one weight via a single parameter apply to false negatives. Setting this weight to the ratio of RWWCE's marginal false negative cost and marginal false positive cost creates equivalent behavior.

For binary classification, the main contribution of RWWCE is a framework to decide its two weights. Starting with the principle that training of a neural network should use the loss function that represents the goal of the end user, RWWCE's two weights, the marginal cost of a false negative and marginal cost of a false positive, should be set at the estimated real world values. For example, if the marginal cost of missing a disease (a false negative) is $2,000 in estimated future medical care and pain and suffering, and a false positive costs $100 in unnecessary testing, the RWWCE's marginal false negative weight and marginal false positive weight should be set at 2000 and 100 respectively. In the case of imbalanced classes, false negatives are often the bigger issue. However, it could also be the case that a false positive has a high cost - perhaps a false positive triggers expensive, unnecessary treatments and loss of confidence in the test. The estimation of these values should come from domain experts. The RWWCE's weights should be set once and only adjusted when domain experts believe they have better estimates, or the world has changed. These weights are not hyperparameters of the neural network for machine learning practitioners to constantly tune (such as the number of layers, optimizer, or learning rate).

Oversampling and undersampling are also heuristics applied to imbalanced classes. Oversampling introduces additional computational complexity better solved with adding weights on the less frequent class (either with out of the box weighted binary cross-entropy loss function or RWWCE loss function). Undersampling loses information from the training set. And setting the degree of undersampling while maintaining sufficient data becomes yet another hyperparameter to estimate and tune.

## V. FALSE NEGATIVES IN SINGLE-LABEL, MULTICLASS CLASSIFICATION

We tested the RWWCE in a scenario designed to demonstrate a high cost mistake in a single-label, multiclass classification problem. Examples of this type of mistake include misclassifications considered racist, and expensive diagnostic error in a medical context.

We again used the MNIST data set. We again divided up our data set into 67.5% training, 7.5% validation, and 25% test.

We trained 90 control neural networks and 90 experimental neural networks. Their only difference was loss function. All were trained with 10 epochs and a batch size of 100 with the Adam optimizer. All took 784 inputs, then a dense layer of 50 and a penultimate dense layer of 20. All layers except the final layer were activated with ReLU. The final layer was a dense layer of 10, activated with softmax, representing the prediction for each of 10 possible classes (e.g. odds of a digit "0", "1", "2",…, "9"). Softmax is the standard activation for single-label, multiclass classification. All were implemented with the same Keras library using TensorFlow as the backend.

For each pair of control and experimental model, the pair was created, trained, and tested together for a total of 90 runs. Each of the 90 runs used a different high-cost combination of false negative and false positive labels, e.g., (1, 2) means that mislabeling an image of a "1" as a "2" would incur extra penalty. Within a run, the control and experimental models were analyzed on the same pair of false negative and false positive. There are 90 possible combinations of (false negative, false positive), so the 90 experimental neural networks represent all possible mislabeling mistakes.

The 90 experimental neural networks were trained using the RWWCE loss function. For each, the specific high cost combination of false negative and false positive was set at a cost of 19, and the other false negative costs left at 1. This represents a domain expert estimating the marginal cost of a false negative in most cases at $1, but the marginal cost of a specific false negative, false positive pair at $20, perhaps due to the social cost of reinforcing a social bias or mistakenly predicting a lower acuity disease when a higher acuity disease is present.

The results demonstrate a reduction in the number of mislabeled digits as well as Real World Cost. There is an increase in top-1 error, which is expected. The neural network training step no longer maximizes overall accuracy but rather minimizes Real World Cost, which is reflected in the reduction in high cost mistakes. The p-values are calculated using the pairwise t-test.

TABLE II
RESULTS OF RWWCE LOSS FUNCTION FOR
SINGLE-LABEL, MULTICLASS CLASSIFICATION

| Model | Mean number of high cost (FN,FP) | Top-1 Error | Real World Cost |
|---|---|---|---|
| Control (n=90) | 6.67 | 3.56% | $0.0428 |
| Experimental (n=90) | 2.57 | 3.62% | $0.0390 |
| p-value | $1.2 \times 10^{-11}$ | $1.1 \times 10^{-2}$ | $4.5 \times 10^{-8}$ |

## VI. SOFTMAX

We review the design of softmax [19] and its associated loss function, categorical cross-entropy loss function from Equation (3). When activating a layer with softmax, the outputs sum to 1 and the interpretation is that each of the k outputs estimates the probability or confidence that the class is present. For example, suppose there are three classes of images: dogs, trees, and everything else. An output of [0.6, 0.3, 0.1] is interpreted to mean there is a 60% chance of a dog, 30% chance of a tree, and a 10% chance of something else. However, the categorical cross-entropy loss function only penalizes the 40% confidence that the image does not show a dog (a probabilistic false negative). It does not penalize the 30% chance of a tree (a probabilistic false positive).

An argument in favor of this design is that because the negative log loss function is concave, a weighting of [0.6, 0.3, 0.1] has a higher loss than [0.6, 0.2, 0.2], despite both results being identical. Therefore, the probabilistic false negatives should be ignored so that the losses are identical. Essentially, the loss function says, "I don't care what softmax predicts on the negative classes, I only care that it predicts 60% on the positive class". The consequence of his design is that the typical loss function applied to softmax is unable to directly penalize probabilistic false positives. The RWWCE was designed specifically to provide the ability to increase the penalty of specific probabilistic false negatives for specific target labels.

Softmax is a generalization of binary classification, however, the implementation details differ. In binary classification, there is typically only one output that predicts the probability of a positive. There is not a second output that predicts the probability of a negative. This means that the binary cross-entropy loss function, Equation (1), must impute negatives via the right hand side of the expression being summed:

$$J_{bce} = -\frac{1}{M}\sum_{m=1}^{M}[y_m \times \log(h_\theta(x_m)) + (1-y_m) \times \log(1-h_\theta(x_m))]$$

In recent work that showed an unexpected benefit of the categorical cross-entropy function, Mahajan *et al.* recently found that it worked better than binary cross-entropy in the case of a multilabel problem [20]. This was unexpected because binary cross-entropy is the theoretically preferred approach for multilabel categorization, such as when images could include both a cat and a dog. Our observation is that each class was very imbalanced, with most images labeled with two hashtags (i.e. labels) among thousands of possible labels. For any single label, a binary cross-entropy loss function would suffer from the classic problem of imbalanced classes: there would be far more true negatives than true positives. The classifier could achieve high accuracy by simply never predicting any hashtag, directly analogous to our initial thought experiment. Mahajan *et al.* found that applying the categorical cross-entropy loss function against only the limited number of positive target labels performed better, which we believe is the result of the categorical cross-entropy loss function focusing on penalizing probabilistic false negatives rather than probabilistic false positives during training. In future work, we will test the performance of the binary RWWCE loss function against the results of Mahajan *et al.*

## VII. CONNECTION TO MAXIMUM LIKELIHOOD ESTIMATION

Suppose a binary classification scenario in which incorrectly missing a disease costs $100 of future healthcare expenditure and patient pain and suffering, and a false positive costs $5 in retesting and loss of credibility. Intuitively, the false negative should be weighted 20 times more: one false negative costs the same as 20 false positives. During neural network training, the binary RWWCE loss function is given by Equation (5):

$$J_{brwwce} = -\frac{1}{M}\sum_{m=1}^{M}[100 \times y_m \times \log(h_\theta(x_m)) + 5 \times (1-y_m) \times \log(1-h_\theta(x_m))]$$

This can be rewritten as:

$$= -\frac{1}{M}\sum_{m=1}^{M}[y_m \times \log(h_\theta^{100}(x_m)) + (1-y_m) \times \log((1-h_\theta(x_m))^5)]$$

Because the exp and log functions are monotonic, $y$ only takes on the values of zero or one, and the $1/M$ factor is a constant; minimizing for the above is equivalent to maximizing for:

$$= \prod_{m=1}^{M} y_m \times h_\theta^{100}(x_m) + (1-y_m) \times (1-h_\theta(x_m))^5$$

The $y_m$ and $1 - y_m$ terms are essentially a conditional statement embedded in an equation. The $h_\theta(x_m)$ and $1 - h_\theta(x_m)$ terms comprise the probability of predicting the target. Based on those terms, we can easily define a probability function $f(y|x)$, the probability of a machine learning model predicting $y$ given inputs $x$. We also add the condition theta to make explicit the trainable weights of the neural network. We then express the above as:

$$= \prod_{m=1}^{M} f^X(y_m \mid x_m, \theta) \quad (7)$$

where $X = 100$ if $y$ is one and $X = 5$ if y is zero.

Under the assumption of an independently drawn and identically distributed training set (i.i.d.), this is also the joint probability. In future work, we will explore whether the i.i.d. assumption indeed holds.

$$= f(y_1, \ldots, y_1, y_2, \ldots, y_2, \ldots, y_m, \ldots, y_m \mid x_1, \ldots, x_m, \theta)$$

where $y_m$ is repeated 100 times if equal to 1 or repeated 5 times if equal to 0.

We have just tied our loss function backwards towards the form of Maximum Likelihood Estimation, the principle that the parameters theta (weights of a neural network machine learning model) that have the highest *probability* of predicting the observed data also have the *maximum likelihood* of being the correct thetas given the observed data. As is conventional, we reverse the observations and the thetas vector to make explicit that we are searching over the space of theta based on fixed observations to maximize likelihood of the correct thetas.

$$= \mathcal{L}(\theta | y_1, \ldots, y_1, y_2, \ldots, y_2, \ldots, y_m, \ldots, y_m, x_1, \ldots, x_m)$$

where $y_m$ is repeated 100 times if equal to 1 or repeated 5 times if equal to 0. Effectively, each time our training data has target 1, we are asking to optimize thetas as if it has seen 100 occurrences (or if target zero, five occurrences). In fact, Equation (7) is the form of the weighted likelihood estimator of Hu and Zidek as seen in [8, Def. 2, Ex. 2].

An example is to consider the simple problem of a Bernoulli Trial with real world costs.

Suppose a game is played at an amusement park: predict a coin flip. When a head is correctly predicted, the observer wins 9 stickers. Tails, she wins 1 toy. A wrong prediction wins nothing. At the end of the game, what will be the ratio of stickers to toys?

The observer builds a binary classifier that takes no inputs and outputs a single value, p, which is typically interpreted as the probability of heads but as we will soon see may have a better interpretation. This value is both the output and the sole trainable weight of the neural network (or theta value). The RWWCE loss function is used with the cost of a marginal false negative at 9 and marginal cost of false positive at 1. Again, a false negative is a prediction of tails when the outcome is heads, losing the opportunity to win 9 stickers. A false positive is predicting heads when the outcome is tails. We train on 2 randomized examples, with 1 head and 1 tail.

The cost function is plotted below. The goal of gradient descent while training the machine learning model is to approximate the minimum located near $p=0.9$. We prove that the minimum is indeed $p=0.9$. From Equation (1), where $p = h_\theta(x_m)$, the loss is given by:

$$J_{bce} = -\frac{1}{10}[\log(p) + \log(p) + \log(p) + \log(p) + \log(p) \\ + \log(p) + \log(p) + \log(p) + \log(p) \\ + \log(1-p)]$$

The minimum is found where the first derivative is zero. We eliminate the -1/10 term.

$$0 = \frac{dJ_{bce}}{dp}$$

$$0 = [\log(p) \times 9 + \log(1-p)]'$$

$$0 = \frac{9}{p} - \frac{1}{1-p}$$

$$\frac{9}{p} = \frac{1}{1-p}$$

$$\frac{p}{9} = \frac{1-p}{1}$$

$$p = 9 - 9p$$

$$10p = 9$$

$$p = 0.9$$

This number can be used to correctly predict the ratio of stickers to toys (observe that the 1-p equals 0.1, so the ratio is 0.9:0.1 or 9:1).

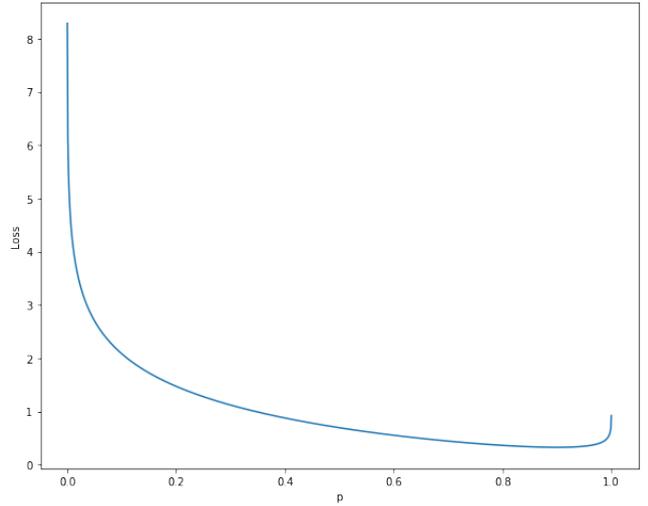

**FIGURE 1.** Value of loss function J at various values of p for scenario

To confirm the identicalness of gradient descent to weighted maximum likelihood estimation, the task is to calculate the parameter p that maximizes the joint probability of predicting the output (1,1,1,1,1,1,1,1,1,0). Equation (7) gives

$$\mathcal{L}(\theta|1,1,1,1,1,1,1,1,1,0)$$

$$= f(1,1,1,1,1,1,1,1,1,0|\theta)$$

$$= p \times p \times p \times p \times p \times p \times p \times p \times p \times (1-p)$$

$$= p^9 - p^{10}$$

The maximum value of $\mathcal{L}$ is found where the first derivative is zero:

$$0 = 9p^8 - 10p^9$$
$$9 = 10p$$
$$0.9 = p$$

The equivalence of the results demonstrates that for this example, gradient descent with a binary RWWCE loss function is equivalent to Maximum Likelihood Estimation on imputed observations of real world outcomes (stickers and toys), which are imputed from actual observations (heads and tails) and knowledge about how those actual observations drive real world outcomes.

This leads us to our intuition on why the RWWCE works. As previously mentioned, the value p is obviously is not the odds of a head, which is still 0.5. Our interpretation is that the value p is a prediction about real world outcomes, namely, the odds of receiving a sticker (0.9) versus a toy (0.1) as the next prize. We believe that the RWWCE cost function is allowing direct optimization of real world outcomes. It does so in a manner that is equivalent to imputing observations of real world outcomes and then applying Maximum Likelihood Estimation on those imputed observations.

In a sense, the Real World Cost metric is a transformation of the observed values (heads or tails or disease present) to relevant imputed real world values (stickers or toys or dollars). The RWWCE then optimizes for the imputed real world values.

So far, we have only analyzed Maximum Likelihood Estimation in the case of a binary classifier that does not take any inputs. In future work, we will analyze Maximum Likelihood Estimation in the context of single-label, multiclass classifiers that are conditional on inputs.

**ACKNOWLEDGMENT**

We acknowledge an unknown person. In 2016, a user with the id ayalalazoro correctly posed the problem of how to differentially weight false negatives and false positives by class on the Keras github website [21].

We also acknowledge Israel Niezen and Bjorn Eriksson for encouraging research into this area of machine learning; Chiara Cerini for her guidance on statistics and research; and Andrew Ng, Geoffrey Hinton, and Ian Goodfellow for their online courses and books, which we referred to as we explored the mathematical underpinnings of machine learning.

**REFERENCES**

[1] N. Japkowicz, S. Stephen, "The class imbalance problem: a systematic study," *Intell. Data Anal.*, vol. 6, no. 5, pp. 429–449, Nov. 2002, DOI: 10.3233/IDA-2002-6504.

[2] T. Hasanin, T.M. Khoshgoftaar, J.L. Leevy, N. Seliya, "Examining characteristics of predictive models with imbalanced big data," *Journal of Big Data*, vol. 6, no. 69, Jul. 2019, DOI: 10.1186/s40537-019-0231-2.

[3] J. Zou, L. Schiebinger, "AI can be sexist and racist — it's time to make it fair*,*" *Nature,* vol*.* 559, pp. 324-326, Jul. 2018, DOI: 10.1038/d41586-018-05707-8.

[4] M. Graber, R. Wachter, C. Cassel, "Bringing diagnosis into the quality and safety equations," *Journal of the Amer. Med. Assoc.*, vol. 308, no. 12, pp. 1211-1212, Sep. 2012, DOI: 10.1001/2012.jama.11913.

[5] M. Abadi, A. Agarwal, P. Barham, *et al*., "TensorFlow: Large-scale machine learning on heterogeneous systems," 2015. [Online]. Available: http://download.tensorflow.org/paper/whitepaper2015.pdf

[6] Y Cui, M. Jia, T. Lin, Y. Song, S Belongie, "Class-Balanced Loss Based on Effective Number of Samples," presented at *Conf. on Comput. Vision And Pattern Recognit.*, Long Beach, CA, USA, 2019.

[7] T. Lin, P. Goyal, R. Girshick, K. He, P. Dollar, "Focal Loss for Dense Object Detection," in *2017 IEEE Int. Conf. on Comput. Vision (ICCV)*, Venice, Italy, 2017. DOI: 10.1109/ICCV.2017.324.

[8] M.Z. Kukar and I. Kononenko, "Cost-Sensitive Learning with Neural Networks," in *Proc. of the Eur. Conf. on Artif. Intell.*, Brighton, UK, 1998, pp 445-449.

[9] H. He, E.A. Garcia, "Learning from imbalanced data," *IEEE Trans. Knowl. and Data Eng.*, vol. 21, no. 9, pp. 1263–1284, Sep. 2009, DOI: 10.1109/TKDE.2008.239.

[10] N. V. Chawla, K. W. Bowyer, L. O. Hall, W. P. Kegelmeyer, "SMOTE: Synthetic Minority Over-sampling Technique," *Journal of Artif. Intell. Res.*, vol. 16, pp. 321-357, Jun. 2002, DOI: 10.1613/jair.953.

[11] T. Bolukbasi, K. Chang, J. Zou, V. Saligrama, and A. Kalai, "Man is to computer programmer as woman is to homemaker? debiasing word embeddings," in *Conf. on Neural Inf. Processing Syst. (NIPS)*, Barcelona, Spain, 2016, pp. 4349–4357. Available: http://papers.nips.cc/paper/6227-man-is-to-computer-programmer-as-woman-is-to-homemaker-debiasing-word-embeddings

[12] P. Thomas, B. Castro da Silva, A. Barto, S. Giguere, Y. Brun, E. Brunskill, "Preventing undesirable behavior of intelligent machines," *Science,* vol. 366, no. 6468, pp. 999-1004, Nov. 2019, DOI: 10.1126/science.aag3311.

[13] R. Madonik, "When It Comes to Gorillas, Google Photos Remains Blind," *Wired Magazine,* 2018*.* Available: https://www.wired.com/story/when-it-comes-to-gorillas-google-photos-remains-blind/

[14] F. Hu, J. Zidek, "The weighted likelihood," *The Can. Journal of Statist.*, vol. 30, no. 3, Sep. 2008, DOI: 10.2307/3316141

[15] I. Goodfellow, Y. Bengio, I. Courville, "Machine Learning Basics," in Deep Learning, MIT Press, 2016, pp. 131–135. Available: http://www.deeplearningbook.org

[16] G. Dimitroff, L.Tolosi, B. Popov, G. Georgiev, "Weighted maximum likelihood as a convenient shortcut to optimize the F-measure of maximum entropy classifiers," in *Proc. of Recent Advances in Natural Lang. Process.*, Hissar, Bulgaria, 2013. Available: https://www.aclweb.org/anthology/R13-1027.pdf

[17] E. Eban, M. Schain, A. Mackey, A. Gordon, R. Rifkin, G. Elidan, "Scalable learning of non-decomposable objectives," in *Proc. of the 20th Int. Conf. on Artific. Intell. and Statist.*, Fort Lauderdale, FL, USA, Apr. 20-22, 2017.

[18] A. Ng, "Handling Skewed Data," in *Machine Learning*, Coursera, 2012.

[19] J.S. Bridle, "Probabilistic Interpretation of Feedforward Classification Network Outputs, with Relationships to Statistical Pattern Recognition," *Neurocomputing, NATO ASI Series*, vol. F 68, 1990. DOI: 10.1007/978-3-642-76153-9_28.

[20] D. Mahajan, R. Girshick, V. Ramanathan, K. He, M. Paluri, Y. Li, A. Bharambe, and L. van der Maaten, "Exploring the limits of weakly supervised pretraining," in *Eur. Conf. on Comput. Vision (ECCV)*, Munich, Germany, Oct. 2018, pp. 181–196. DOI: 10.1007/978-3-030-01216-8_12.

[21] Ayalalazaro, "Is there a way in Keras to apply different weights to a cost function?," 2016. Available: https://github.com/keras-team/keras/issues/2115